\documentclass[11pt,a4paper]{article}
\usepackage[hyperref]{acl2020}
\usepackage{times}
\usepackage{xcolor}
\usepackage{censor}
\usepackage{natbib}
\usepackage{latexsym}
\usepackage{float}
\usepackage{booktabs}
\usepackage{xcolor}

\usepackage{microtype}
\usepackage{verbatimbox}
\usepackage{tikz}
\usetikzlibrary{arrows.meta,
                chains,
                positioning,
                shapes.geometric}
\aclfinalcopy 

\title{Raply: A profanity-mitigated rap generator}

\author{
Omar Manil BENDALI, 
Samir FERROUM,
Ekaterina KOZACHENKO,
Youssef PARVIZ,\\
\textbf{Hanna SHCHARBAKOVA,
Anna TOKAREVA,
Shemair WILLIAMS}
  }

\date{}

\begin{document}
\maketitle
\begin{abstract}

The task of writing rap is challenging and involves producing complex rhyming schemes, yet meaningful lyrics. In this work, we propose \textit{Raply}, a fine-tuned GPT-2 model capable of producing meaningful rhyming text in the style of rap. In addition to its rhyming capabilities, the model is able to generate less offensive content. It was achieved through the fine-tuning the model on a new dataset $Mitislurs$, a profanity-mitigated corpus. We evaluate the output of the model on two criteria: 1) rhyming based on the rhyme density metric; 2) profanity content, using the list of profanities for the English language. To our knowledge, this is the first attempt at profanity mitigation for rap lyrics generation.
\end{abstract}

\section{Introduction}

Since its inception in the late 1980’s rap music has become one of the most influential music genres owing to its ability to address various ethical and social issues affecting people worldwide.

In this paper, we study the problem of determining the most lyrically relevant rap line while taking into account ethical issues. Our interest in this problem is motivated by the following perspectives. Firstly, we want to study the structure of rap lyrics to generate relevant text which adheres to the features of rap lyrics.
Secondly, we are aware of social and ethical issues around this topic, but we also should treat respectfully the specific range of language expressions that is often characteristic of the rap genre. Thirdly, we want to be able to give rappers the ability to create expressive works about topics relevant to their experience without the need to fall back on human ghostwriters when the works they create are not lyrically strong (i.e. lack of rap techniques).

We formalize this problem as an auto-completion task. 
Given a query $p = [A:, s (1), s(2), ..., s(n) ]$, an auto completion returns $q = [B:, t(1), t(2) , ..., t(n) ]$, where q is an extension of p. In this specific rap lyrics generation problem,  $p$ is the set of the input tokens $s(i)$, and $q$ is the set of model's generated tokens $t(n)$ that shares syntactic, semantic and rap-specific characteristics with $p$ (rhyming), we assume that $A$ marks the beginning of the user input line and $B$, the beginning of the model's generated line.

Our main contributions can be summarized as follows:
\begin{enumerate}
\item We propose a method for mitigating the number of profanities in rap lyrics generation through the construction of the $Mitislurs$ corpus, in order to be mindful of the issues often discussed in the rap genre.

\item We introduce a method of rap lyrics generation that is able to produce satisfying rap in terms of rhyming.
\end{enumerate}

\textbf{Content Warning.} This paper contains material that may be offensive or upsetting. In contrast to the Appendix \ref{sec:appendix}, where the abusive language examples are provided verbatim, examples in the main text are represented by initial letters of the profanities followed by asterisks. The authors oppose the use of abusive language.

\section{Related work}
Automatic rap lyrics generation is a challenging task in natural language processing. It requires a system to be able to learn complex linguistic and semantic information from lyrics and to use this knowledge to create a new, creative song.

This task has become a more active research area in recent years, mainly due to the success of deep learning techniques \citep{potash-etal-2015-ghostwriter, malmi-etal-2016-dopelearning, manjavacas-etal-2019-generation, nikolov-etal-2020-rapformer, liu-etal-2022-chipsong}. 

\cite{potash-etal-2015-ghostwriter} show the effectiveness of an LSTM model for rap lyrics generation: using its own rhyme scheme, line and verse length, \textit{Ghostwriter} can generate lyrics similar to the rap style of a specific rapper. In \cite{malmi-etal-2016-dopelearning} the authors address the problem of rap lyrics generation as an information retrieval task: the system treats input line as a query to which it predicts the most suitable next line. The Transformer-based \cite{vaswani-etal-2017-attention} approach to the generation of rap using content words from the song is described in \cite{nikolov-etal-2020-rapformer}. \cite{liu-etal-2022-chipsong} proposed a \textit{ChipSong} – a lyrics generation system that enables the user to control different attributions of the generated song such as word-level format control and rhyme control. 

We also considered studies related to poem generation, as they share similar objectives in terms of rhyme production. A study by \cite{lo-etal-2022-gpoet} introduced GPoeT-2 model, based on GPT-2 architecture and capable of generating from any given subject. Lastly, \cite{hamalainen-etal-2022-modern}, also sought to establish a model capable of generating French poems by using two pre-trained language models.

However, to the best of our knowledge, none of the previously mentioned studies take into account the presence of profanity, although based on a study conducted by \cite{frisby-behm-2019-undressing}, among the many musical genres, rap is the one that uses the most profanities compared to other genres. Considering that large language models demonstrate bias and toxicity \citep{gehman-etal-2020-realtoxicityprompts, sheng-etal-2019-woman, wallace-etal-2019-universal}, it can be assumed that the resulting output from the models may contain offensive language.

\section{Rhyming and rap}
Rhyme may be defined as the “acoustic agreement of sounds (vowels and consonants), words or groups of words” \cite{Grofcikova2020-vh}. In other words, rhyme is simply having a similar sound between two or more words. Rhyming is a defining feature of various works of art such as poetry, songwriting as well as rap. 

\subsection{Types of rhymes}
Rhymes may take various forms, such as perfect rhymes or slant rhymes, to name a few. Perfect rhymes are one of the most popular and well-known types of rhyming which involves words sharing the same syllabic sound stress occurring at the end of the word. The sound similarity and syllabic stress between the words \textit{walk} and \textit{talk}, \textit{stalk} are examples of perfect rhymes. While slant rhymes, which is a differentiation of perfect rhymes, do not have such a strict sound pattern, and the rhyme is often located internally within the words. Take for example the words, \textit{home} and \textit{none}, the similarity exists internally within each word as the word stress happens at the beginning of both words resulting in a rhyme. 

Other not-so-common forms of rhyme include assonance which involves the repetition of vowels for example, \textit{drown} and \textit{sound}. Consonance, another form of rhyme is characterized by rhyming through the repetition of consonant sounds such as \textit{staff} and \textit{flat}. Many of these styles of creating rhyme overlap or work together to create rhythm. For instance, alliteration is a combination of assonance and consonance where the same consonant or vowel is repeatedly stressed in each word. The sentence \textit{Peter Piper picked a peck of pickled pepper}, is a perfect example of alliteration at play.

Rhymes may be further defined by their positions within a line or sentence or verse. End rhymes for instance appear at the end of a sentence or line while internal rhymes take place within a line or sentence. 

\subsection{Rap song structure}
Various artists constantly make use of rhyming schemes to create rhythm and rhyming patterns, especially in rap music. Many rappers follow some form of the rhyming scheme in their music and this is quite apparent in the various rap songs chosen for our corpus. And in like manner, these rhyme patterns may be found either at the beginning or the end of a line, sentence, or verse.

Rap music makes use of bars, which is simply another name for the lyrics of the lines the rappers use in their music. These bars are then grouped by what are known as verses. Verses refer to the different segmentations that occur in a rap song. In general, rap music employs about 16 bars per verse. Each verse used by the rapper tells some form of a story and is dependent on some form of rhyming scheme. In addition to this, many verses are preceded by what is known as a hook or the punchline of the verse. This is the part of the verse which captures and maintains the listener's attention throughout the entirety of the lyrics.

\section{Data collection}
In this section, we describe the steps we made to obtain the new corpus and preprocess it.

To build a system that generates a rap line based on the text input, we compiled a large corpus of rap lyrics together with lyrics metadata. For this purpose, we scrapped Genius.com \footnote{https://genius.com/.}, a large collection of lyrics, to obtain the necessary data.

Foremost, we assembled a list of 89 American English-speaking rap performers chosen based on their renowned writing and rhyming skills. Further, via the free Genius Song Lyrics API request, which is made using a client Access Token generated on the site, we got access to all available information about that particular performer. With the help of \textit{lyricsgenius} library \footnote{https://lyricsgenius.readthedocs.io/en/master/.}, we collected the required information, specifically the artist’s name, the name of the song, the year of release of the song, and the lyrics.

Thus we obtained a raw corpus of 21,801 songs with a size of 14,296,438 tokens.

We use preprocessing which was designed with attention to the specifics of the data. Based on the structure of the rap song, we removed parts of the songs that we found unnecessary (e.g., instrumental) and repetitive (e.g., song remixes). In addition, we removed all the songs written in languages other than English. In songs, the parts that represent the chorus of songs are repetitive, and would not help to generate sophisticated lines; also, interludes, do contain unnecessary discourse that would not be relevant to our task. There are also confusing parts, that were dropped, which do not exist in all the songs (e.g., \textit{[Intro]} or non-ASCII characters). Additionally, we have chosen to keep only verses parts. The aim of using verses only is to focus only on parts of the lyrics that contain the most rhyming pieces, other parts are considered repetitive and irrelevant to the rhymes.

\section{Data annotation} \label{section:annotation}
\begin{figure}[h]
    \centering
    \includegraphics[width=0.53\textwidth]{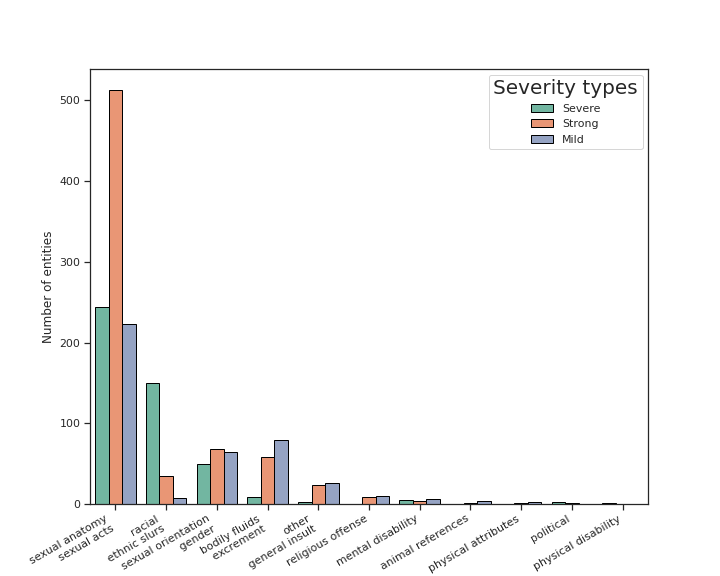}
    \caption{The distribution of profanities by categories according to their severity ratings in \textit{The Obscenity List}.}
    \label{fig:severity}
\end{figure}

Modern language models are trained on large text corpora from the Internet. If a model receives training data containing slurs or abusive language, it will learn to predict these words during the training step and subsequently produce the output containing these words. 
To minimize the occurrence of profanities in the data generated by the system, we decided to automatically filter data by offensive language beforehand.
To compile the dataset, we used \textit{The Obscenity List}, developed by Surge AI \footnote{https://www.surgehq.ai/.}, which contains over 1600 popular English profanities and their variations \footnote{https://github.com/surge-ai/profanity.}, including non-standard spellings (e.g., canonical form \textit{damn} could be written as \textit{g0ddamn}). Figure \ref{fig:severity} illustrates the distribution of profanities by category according to their severity ratings in \textit{The Obscenity List} \footnote{Severity ratings distribution [1-3]: from 1 to 1.4 - mild, strong, and from 1.5 to 2.4 - strong, and from 2.5 to 3 - severe.}. It can be seen that the prevailing majority of profanities are sexual anatomy/ sexual acts, having a strong degree of severity. The second most frequent group is racial/ethnic profanities, the level of severity of which is, for the most part, the maximum.

\begin{figure}[H]
    \centering
     \includegraphics[width=.5\textwidth]{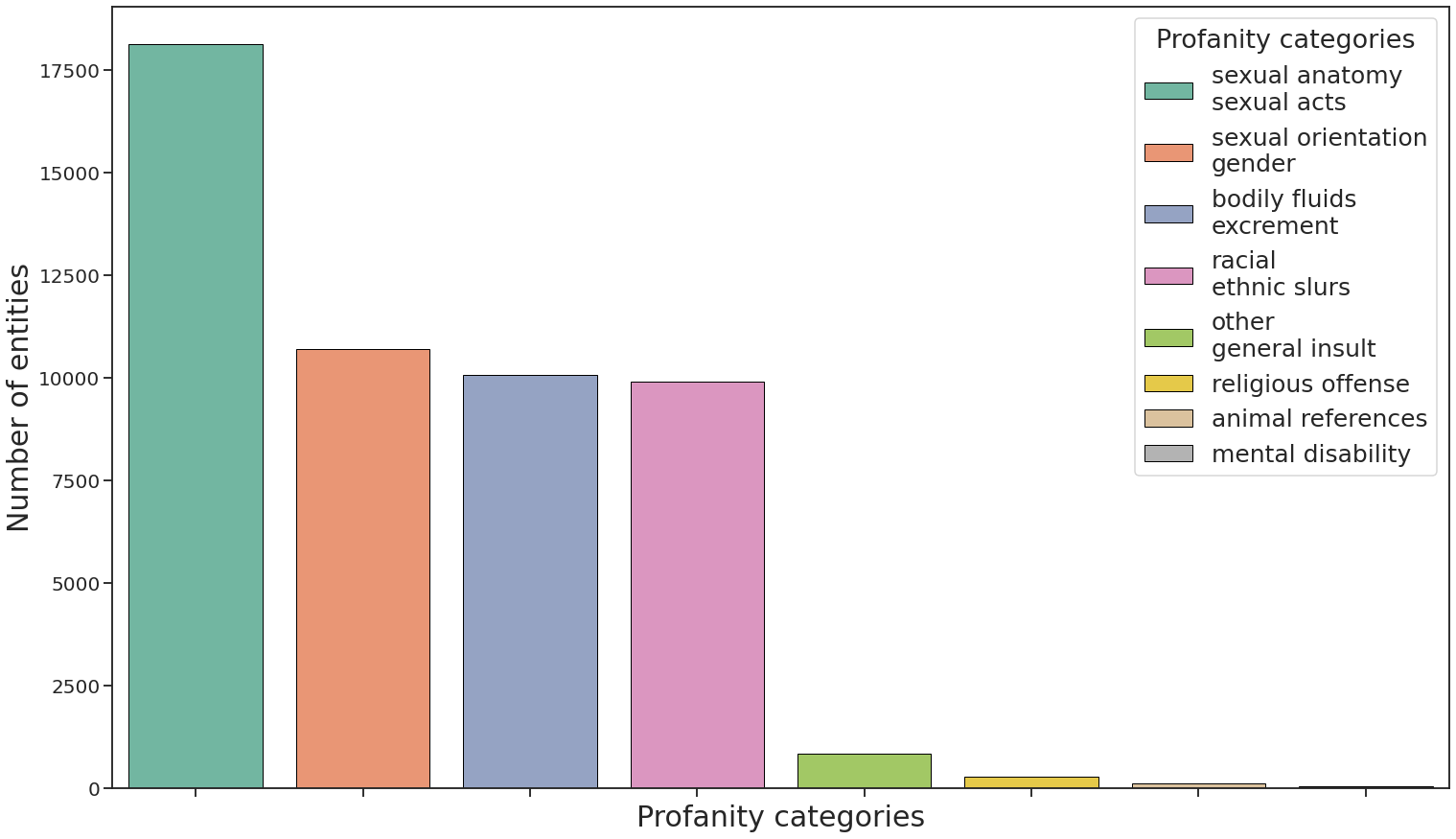}
    \caption{Distribution of profanities across matched categories in the raw dataset.}
    \label{fig:scheme}
\end{figure}

In order to tokenize and lemmatize each lyric from the raw dataset, we used \textit{NLTK} library (\textit{word\_tokenize()} to tokenize the texts and the \textit{Wordnet Lemmatizer} to lemmatize them). Then we automatically extracted the profanities from each song. The obtained dataset with offensive language (see Figure \ref{fig:scheme}) includes lyrics in which 50,130 profanities were found, as well as lyrics with no profanities at all, that is 2,225 songs.

Moreover, we additionally used two types of annotation, which were produced by Surge AI \footnote{In the Appendix \ref{sec:appendix} are shown some examples of the profanities from each group.}: profanity categories and severity rating.

\textbf{Profanity categories.} Firstly, we performed the annotation of each profanity for one of 11 profanity's primary categories: 
\begin{itemize}
\item sexual anatomy / sexual acts 
\item bodily fluids / excrement
\item sexual orientation / gender
\item racial / ethnic slurs
\item mental disability
\item physical disability
\item physical attributes
\item animal references
\item religious offense
\item political
\item other / general insult
\end{itemize}

As can be seen in Figure \ref{fig:scheme}, only 8 out of 11 profanity categories from the \textit{The Obscenity List} were found in a compiled dataset. The most frequent profanity category is sexual anatomy / sexual acts.

\textbf{Severity rating.} In \textit{The Obscenity List}, the severity rating annotation was done manually by 5 Surge AI experts. The data annotators were asked to rate how severe they believed each profanity to be, on a 1-3 point scale. The rating reflected in the dataset is the mean of those 5 ratings. We took this type of annotation in order to get the values of the $slur\_score$ for each lyric, which was calculated using the following formulas:\\

$ws\_score(line) = \frac{\Sigma sev\_ratings(line)}{number\_of\_tokens(line)}$

\begin{center}
    $ slur\_score  =  \frac{\Sigma ws\_score}{number\_of\_lines}$
\end{center}

\begin{figure}[H]
    \centering
     \includegraphics[width=.5\textwidth]{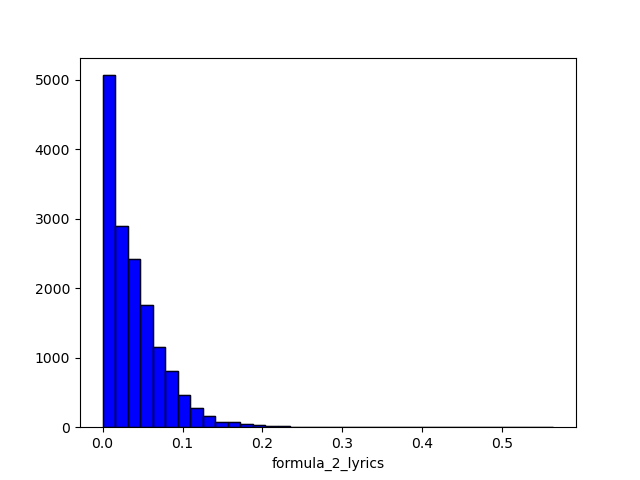}
    \caption{Distribution of slur scores in the raw dataset.}
    \label{fig:scores}
\end{figure}

We first compute the weighted severity score ($ws\_score$) defined as the sum of the severity rating of each profanity identified in one line and average over its number of tokens. Then, we average the sum of weighted severity scores of all the lines of the song over the total number of lines to obtain the $slur\_score$.
Next, we drop all the songs lyrics that have a $slur\_score$ above the threshold value of 0.05 ($3^{rd}$ quantile of the $slur\_score$ distribution, as shown in Figure \ref{fig:scores}). We deliberately choose to filter slurs on a song level only, because of the possibility of breaking lyrical structure and cohesion.

\begin{figure}[H]
    \centering
     \includegraphics[width=.55\textwidth]{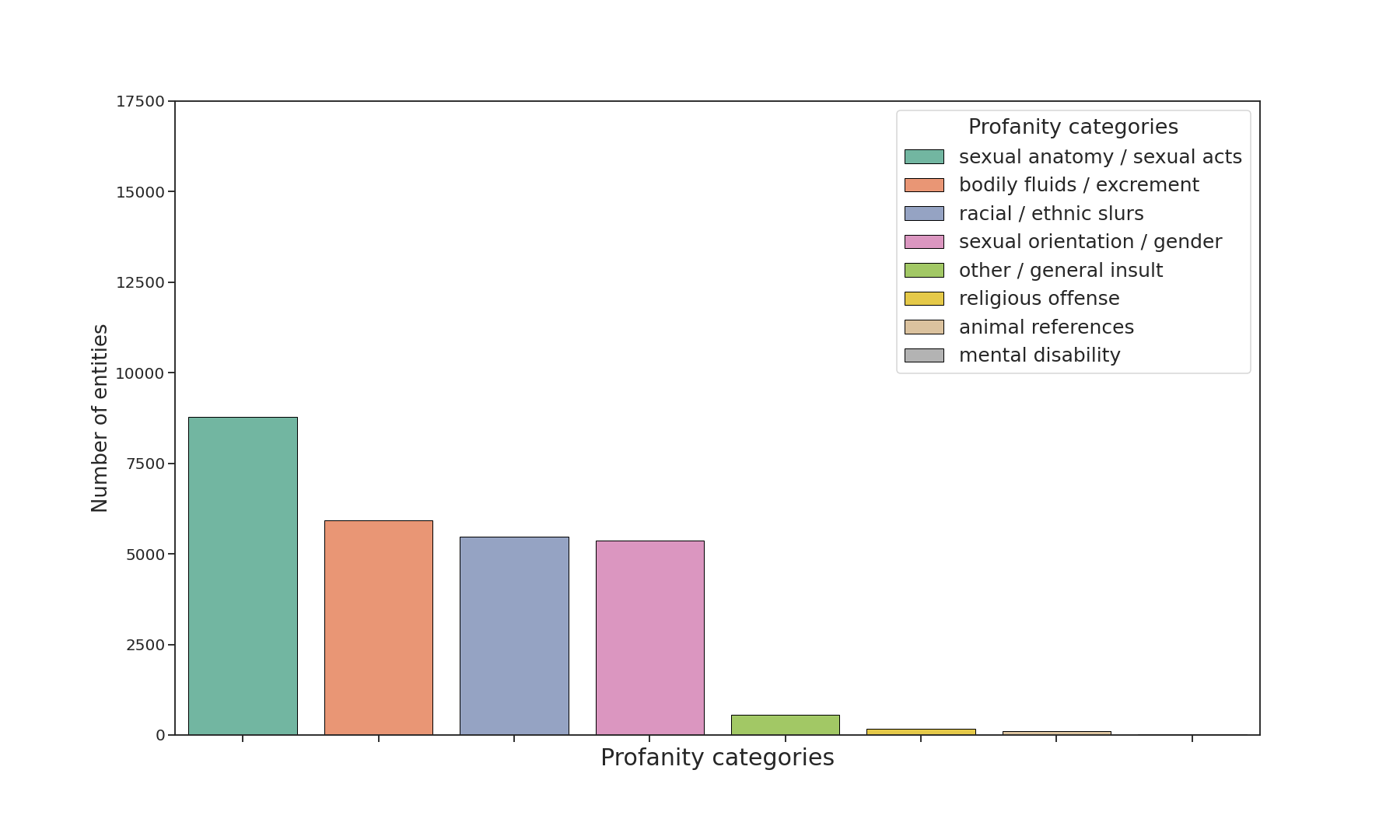}
    \caption{Distribution of profanities across matched categories in the filtered dataset.}
    \label{fig:scheme1}
\end{figure}

As a result of the above steps, for each song lyrics, we extracted profanities and matched them with a list of profanity categories from \textit{The Obscenity List} \footnote{We should note that the set of profanities is limited by \textit{The Obscenity List}.}. The distribution of profanities in the filtered dataset is shown in Figure \ref{fig:scheme1}. 

At the end of these steps, we obtained two corpora:
\begin{itemize}
    \item $Slurs$ corpus, which is the corpus we collected, cleaned, and annotated; it comprises 15,282 songs (6,514,188 tokens).
    \item $Mitislurs$ corpus, composed of 10,748 songs (4,531,673 tokens) which represents a subset of the \textit{Slurs} corpus, that has been filtered to mitigate the occurrence of profanities (as defined in Section \ref{section:annotation}). 
\end{itemize}

\section{Raply}
In this section, we describe \textit{Raply} – an automatic rap generation system.
\subsection{Model Overview}

We used a GPT-2 \cite{radford-etal-2019-language} model, which is a generative language model based on Transformers decoder.
Training such a model comes in two steps: an unsupervised pre-training on a large quantity of data followed by a supervised fine-tuning step. In the case of an auto-completion task, the input of the model is the concatenation of the query $p$ and its extension $q$, with a delimiter token in between. Since our goal is to explore the effectiveness of GPT-2 on generating rap, instead of training the model from scratch, we use a pre-trained GPT-2 model and fine-tune it on our corpora for the task of rap lyrics generation. 
\subsection{Experiments}

We run the experiments on the two built corpora. Due to the nature of input data required for fine-tuning GPT-2 model, we concatenate verse lines, separated by the special token \textit{'Line:'}. The pre-trained model we used was trained on \textit{BookCorpus} \cite{yao-huang-2018-temporal}, a dataset of over 7,000 unpublished fiction books from various genres. We reuse the byte-pair encoding vocabulary of this model.

We randomly select 10\% examples from data as the test set and use the remaining examples as the training set. For both corpora, we use Adam optimizer with a learning rate of $1e^{-5}$ and a batch size of 32. We train the models on 3 epochs. The training phase was done in both Kaggle\footnote{https://www.kaggle.com/.} platform as well as Grid 5000\footnote{https://www.grid5000.fr/.} on Nvidia GPU P100.

After the training phase, we proceed with the generation phase. We split each instance of the test set into two parts, an input part, and a reference part, the input is fed to the model during the generation and the reference is used as a baseline. Our code is made available on GitHub \footnote{https://github.com/ManilBen/Raply.}.

We request the model to generate sequences using \textit{top-$k$ sampling} method \cite{fan-etal-2018-hierarchical}, the value of $k$ is set to 50 and the minimum and maximum length of the output sequence to 4 and 50 respectively.

For each generated sequence, we compute its rhyme density, a metric introduced by \cite{malmi-etal-2016-dopelearning} to evaluate the quality of rhyming fluency.  Rhyme density has shown a correlation with a human evaluation of rap lyrics' technical quality, which makes it a suitable metric for evaluating the models. A rhyme density greater than 1 can be considered as high. It should be taken into account though, that this metric only considers one aspect of rap, namely rhyme, and ignores other crucial components of rap like figurative language, the context, the meaning of the words, etc.

Since rhyming is not always identifiable through orthography, we use the program \textit{eSpeak}\footnote{https://espeak.sourceforge.net/.} to transform each token into its phoneme form in the IPA \footnote{International Phonetic Alphabet.} and then find the longest matching vowel sequence near each token. We then average the length of the sequence overall words to obtain the value of rhyme density. This process is applied to both the generated sequence and the reference part of the test set. We choose to mainly focus on the rhyming aspect. We justify this by the fact that GPT-2 model already showcases satisfying semantic preservation abilities and that rap lyrics are inherently not strict when it comes to semantic preservation in comparison with poetry for example.
We also measure the slur score on the generated text as defined in the formula defined in Section \ref{section:annotation}.

\subsection{Results and discussion}

We compare the performances of the GPT-2 model on our two corpora ($Slurs$ vs $Mitislurs$) and compare the perplexity and rhyme density values of the generated text with the human-generated one from our test set. Perplexity is frequently used to assess the quality of language models, and measures how well a probabilistic model is able to predict a sample. A lower perplexity means that the model is more confident in its predictions and, as a result, is less startled by the data. 

As Table \ref{rd} shows, the model GPT-$2_{Mitislurs}$ is able to improve the rhyme density and outperforms \textit{Ghostwriter} \footnote{The value is adopted from \cite{xue-etal-2021-deeprapper}.} \cite{potash-etal-2015-ghostwriter}. However, we are not able to outperform \textit{Dope Learning} model \cite{dope} in terms of rhyme density. We explain this difference by the fact that \textit{Dope Learning} model does not properly generates new lines from scratch but rather predicts the next line of rap from an already rapper-written sample of lines.
\begin{table}[H]
\centering
\begin{tabular}{p{2.4cm} p{1.6cm} p{1.5cm} p{0.5cm}}
\hline
\textbf{Model} & \textbf{Reference RD} & \textbf{Generated RD} & \textbf{PP}\\ 
\hline
Ghostwriter & - & 0.17 & - \\ 
GPT-$2_{Slurs}$ & 0.6 & 0.7 & 41.6\\
GPT-$2_{Mitislurs}$ & 0.65 & \textbf{0.8} & 46.3 \\
DopeLearning & - & 1.4 & - \\ 
\hline
\end{tabular}
\caption{\label{rd} Comparative results of rhyme density (RD) and perplexity (PP).}
\label{tab:rd}
\end{table}
To assess if the profanity-mitigation approach had an impact on the generated text, we evaluate the slur score of generated text by the developed model while training on both $Slurs$ and $Mitislurs$ corpora, and for a sample from generated data from \cite{dope}, made available by the authors\footnote{\href{https://github.com/ekQ/dopelearning/blob/master/random\_songs.zip}{https://github.com/ekQ/dopelearning/random\_songs.zip.}}

Table \ref{profanity-ratio} shows the average slur score values, we notice that the use of $Mitislurs$ corpus helped lower the occurrences of profanities in the generated text, and outperforms \textit{Dope Learning} model. 
This does not mean that profanities will be completely absent in the generated text, our goal is to mitigate the presence of these words. As Table \ref{tab:lines1} shows a sample from the generated text of GPT-$2_{Mitislurs}$, profanities can still occur.

\begin{table}[H]
\centering
\begin{tabular}{ll}
\hline
\textbf{Model} & \textbf{Slur score} \\
\hline
GPT-$2_{Slurs}$ & 0.02 \\
GPT-$2_{Mitislurs}$ & \textbf{0.008} \\
Dope Learning & 0.23 \\
\hline
\end{tabular}
\caption{\label{profanity-ratio} Obtained values of slur score of GPT-$2_{Slurs}$ and GPT-$2_{Mitislurs}$ models.}
\label{tab:model}
\end{table}

\begin{table}[H]
\begin{tabular}{p{1.2cm} p{5.3cm}}
\hline
\textbf{Input:} & \small And now I'm in the rap game, like the crack \colorbox{brown}{game} \\
\textbf{Output:} & \small I got enemies coming, I play \colorbox{lime}{my} \colorbox{brown}{game} \\
        & \small I can be \colorbox{magenta}{king} or queen, like  \colorbox{lime}{my} style, \\
        & \small I play like the \colorbox{magenta}{King} of Pain "Sonic Yoda" \\
\hline
\textbf{Input:} & \small I take my time when I rhyme lovely, me, nah like ugly n**** \\
\textbf{Output:} & \small \colorbox{pink}{Tapping} out the \colorbox{pink}{trap}  \colorbox{pink}{doors} and holla \colorbox{olive}{'bout me} \\
        & \small B***** they gonna nuke if they can \colorbox{olive}{break me} \\
\hline
\end{tabular}
\caption{\label{miti-sample} Example lines generated by the GPT-$2_{Mitislurs}$ model. Note: the highlighted text represents either rhyme or alliterations.}
\label{tab:lines1}
\end{table}
\begin{table}[H]
\begin{tabular}{p{1.2cm} p{5.3cm}}
\hline
\textbf{Input:} & \small Takin orders from a snitch that sleeps with cops \\
\textbf{Output:} & \small You n***** out to f***, my f*****, you a b**** and you some s*** yo \\
\hline
\textbf{Input:} & \small Tell me that they love me, I don’t give a f**** \\
\textbf{Output:} & \small I wanna put on a black man’s d*** and let y’all f*** out his m********* \\
\hline
\end{tabular}
\caption{\label{slurs-sample} Example lines generated by the GPT-$2_{Slurs}$ model.}
\label{tab:lines2}
\end{table}

\section{Further work}
Pertaining to the direction of this paper, we aim to improve text generation by implementing a ranking algorithm during the lyrics generation phase. Therefore, we should be able to select the line which adheres the most to a rhyming pattern. In order to achieve the aforementioned task, we also aim to train the model on a curated corpus that follows better rhyming composition. That is to say, the source data should have a clearer rhyming pattern that will be able to train the model to produce the desired text output. Additionally, the text generation will be tested using more efficient models like GPT-3 and compared to previously generated data to compare and measure the results obtained from previous models. Lastly, we plan to focus on qualitative analysis of model results.

\section{Conclusion}
In this paper, we present \textit{Raply}, a GPT-2-based model for rap generation, which takes into account offensive content and tries to mitigate its occurrence. We achieved it through the collection of a more ethical corpus $Mitilslurs$ and thus we reduced the presence of offensive content in the generated rap. The developed model is also able to generate close-to-human-like rhyming patterns according to the rhyme density metric. 

To our knowledge, \textit{Raply} is the first model that tackles the issue of offensive content in the context of rap lyrics generation.

\section{Environmental impact}
We do care about the environmental effects of our NLP models on energy consumption and carbon emissions. Below we provide the information considering our models' training. Assuming that in order to train each of the models we applied the maximum GPU power, i.e. 250 watts, then the energy consumed can be obtained using the following formula:
\begin{center}
$E = Pt$
\end{center}

We have therefore calculated the energy consumption for each trained model per day (Table \ref{tab:energy}). In total, therefore, we spent 3,25 kilowatts hour per day on training the models. For comparison, we have borrowed the energy consumption statistics for 2019 from the EIA\footnote{https://www.eia.gov/.} website. Thus, each resident in France consumes 17,6 kilowatts of energy per day, which is 5 times as much as our trained models. 

\begin{table}[H]
\centering
\begin{tabular}{p{2.5cm}cp{1.2cm}p{1.1cm}}
\hline
\textbf{Model} &  \textbf{Training hrs/Day} & \textbf{kWh/Day} \\
\hline
GPT-$2_{Slurs}$ & 6 & 1.5 \\
GPT-$2_{Mitislurs}$ & 7 & 1.75 \\
\hline
\hline
\textbf{Total}  & 13 & 3.25 \\
\end{tabular}
\caption{\label{font-table} Energy consumption of GPT-$2_{Slurs}$ and GPT-$2_{Mitislurs}$ per day.}
\label{tab:energy}
\end{table}

\section{Ethical considerations}
Working with text generation models is challenging in the sense that the trained model is a kind of "black box" which is difficult to supervise. This raises significant ethical issues that must be addressed. As we deal with rap lyrics generation, we anyway encounter ethical issues, in particular, discrimination, as the focus of our paper is shifted to handling profanity categories. This paper contains examples of words that may be abusive to certain categories of people. Unfortunately, the occurrence of slurs in rap lyrics is virtually inevitable, since the presence of profanities is part of rap culture. However, in this paper, we aim to reduce the number of slurs in the generated data. It is also essential to stress that the examples of generated lines containing slurs by no means reflect the attitude of the authors toward any categories of people. There is no way we can affect models already trained on enormous amounts of data, but we attempt to retrain the model on our own data to somehow mitigate occurring slurs and obtain less offensive verses.

Another critical concern that we have to deal with is the bias of our model. Since the Obscenity list consists of unbalanced categories of slurs, and certain categories of slurs (e.g. sexual anatomy/acts) tend to occur in the training data, there is a risk that the model will generate more slurs of the same category. Consequently, one should attempt to mitigate the bias, for example, by balancing the songs in the training lyrics dataset according to the presence of profanities of the particular category, or by trying to control the number of slurs directly during generation.

A further matter of no small significance is that of legal issues. The possibility of user-generated content on websites like Genius violating intellectual property rights (IPR) also needs to be addressed, as Genius disclaims any responsibility for any content posted by users. Because it could be difficult or even impossible for others to utilize the data or build upon it if it is not legally permitted to do so, this can pose issues for reproducibility. It is crucial that Genius, or any other platform, take proactive steps to address these issues and establish a clear, consistent policy for handling user-generated content that can violate IPR.

\newpage

\bibliography{anthology,acl2020}
\bibliographystyle{acl_natbib}

\newpage

\appendix

\section{Appendix}
\label{sec:appendix}

In Table~\ref{tab:appendix} are shown some examples of offensive language related to each profanity group: 
\begin{table}[H]
\begin{tabular}{ll}
\hline
\textbf{Profanity group} & \textbf{Examples} \\
\hline
sexual anatomy / sexual acts & ass kisser, dick, pigfucker \\
bodily fluids / excrement & shit, cum \\
sexual orientation / gender & faggot, tranny,
bitch, whore \\
racial / ethnic & chink, n3gro \\
mental disability & retard, dumbass \\
physical disability & quadriplegic bitch \\ 
physical attributes & fatass, ugly whore \\
animal references & pigfucker, jackass \\
religious offense & goddamn \\
political & China virus \\
other / general insult & mofucker \\
\hline
\end{tabular}
\caption{
The examples of profanities for each group from \textit{The Obscenity List}.
}
\label{tab:appendix}

\begin{table}[H]
\begin{tabular}{p{1.2cm} p{5.3cm}}
\hline
\textbf{Input:} & \small Sicken n*****, what I nickname my gun? \colorbox{brown}{Robitussin} \\
\textbf{Output:} & \small  \colorbox{yellow}{I ain't} with the \colorbox{brown}{firin’} shit, \colorbox{yellow}{I ain't} with the \colorbox{brown}{tusslin'} \\
\hline
\textbf{Input:} & \small Military hat, fatigue pants saggin', Henny'd up \\
\textbf{Output:} & \small Gotta play nice when it doesn't rhyme \\
\hline
\textbf{Input:} & \small Some spacious, others is \colorbox{cyan}{basic}  \\
\textbf{Output:} & \small It's \colorbox{magenta}{good} we move with \colorbox{magenta}{good} \colorbox{cyan}{habit} \\
\hline
\textbf{Input:} & \small All my shit is Bohchah Fam   \\
\textbf{Output:} & \small Oh, wait, wait! I got a n**** on tape \\
\hline
\textbf{Input:} & \small I take my time when I rhyme lovely, me, nah like ugly n**** \\
\textbf{Output:} & \small \colorbox{pink}{Tapping} out the \colorbox{pink}{trap}  \colorbox{pink}{doors} and holla \colorbox{olive}{'bout me} \\
        & \small B***** they gonna nuke if they can \colorbox{olive}{break me} \\
\hline
\\
\textbf{Input:} & \small One of the livest n***** \colorbox{green}{in} New York \\
\textbf{Output:} & \small I've got it \colorbox{green}{in} their \colorbox{pink}{veins}, his rap verse \colorbox{pink}{proclaims} \\
\hline

\end{tabular}
\caption{\label{tab:apxlines} Example lines generated by the GPT-$2_{Slurs}$ model.}
\label{tab:apxlines}
\end{table}
\end{table}


\end{document}